\documentclass{article}
\usepackage{spconf,amsmath,graphicx,amsfonts}
\usepackage{tikz}
\usepackage{bm}
\usepackage{natbib}
\usepackage[colorlinks=True,citecolor=blue,linkcolor=blue]{hyperref}
\usetikzlibrary{arrows}
\usetikzlibrary{calc}

\title{End-to-end attention-based large vocabulary speech recognition}
\name{Dzmitry Bahdanau$^\ast$, Jan Chorowski$^\dagger$, Dmitriy
Serdyuk$^\ddagger$, Phil\'emon Brakel$^\ddagger$ and
Yoshua Bengio$^\ddagger{}^1$}
\address{
  $^\ast$Jacobs University Bremen\\
  $^\dagger$University of Wroc\l{}aw\\
  $^\ddagger$ Universit\'e de Montr\'eal\\
  ${^1}$ CIFAR Fellow}
\begin{document}
%\ninept
%
\maketitle
\begin{abstract}
  Many of the current state-of-the-art Large Vocabulary Continuous
  Speech Recognition Systems (LVCSR) are hybrids of neural networks and Hidden
  Markov Models (HMMs).  Most of these systems contain separate
  components that deal with the acoustic modelling, language modelling
  and sequence decoding. We investigate a more direct
  approach in which the HMM is replaced with a Recurrent Neural
  Network (RNN) that performs sequence prediction directly at the
  character level. Alignment between the input features and the
  desired character sequence is learned automatically by an attention
  mechanism built into the RNN. For each predicted character, the
  attention mechanism scans the input sequence and chooses relevant
  frames. We propose two methods to speed up this operation: limiting
  the scan to a subset of most promising frames and pooling over time
  the information contained in neighboring frames, thereby reducing
  source sequence length. Integrating an n-gram
  language model into the decoding process yields recognition
  accuracies similar to other HMM-free RNN-based approaches.
\end{abstract}
\section{Introduction}
\label{sec:intro}

Deep neural networks have become popular acoustic models for
state-of-the-art large vocabulary speech recognition systems
\citep{hinton2012deep}. 
However, in these systems most of the other components, such as Hidden Markov
Models (HMMs), Gaussian Mixture Models (GMMs) and $n$-gram
language models, are the same as in their predecessors.
These combinations of neural networks and statistical models are often referred to as \emph{hybrid systems}.
In a typical hybrid system, a deep neural network
is trained to replace the Gaussian Mixture Model (GMM) emission distribution of
an HMM by predicting for each input frame the most likely HMM state. These state
labels are obtained from a
trained GMM-HMM system that has been used to perform forced alignment. 
In other words, a two-stage training process is required, in which the older GMM
approach is still used as a starting point.
An obvious downside of this hybrid approach is that the acoustic model is not
directly trained to minimize the final objective of interest.
Our aim was to investigate neural LVCSR models that can be trained with a more
direct approach by replacing the HMMs with a Attention-based Recurrent Sequence
Generators (ARSG) such that they can be trained end-to-end for sequence prediction.

Recently, some work on end-to-end neural network LVCSR systems has shown
promising results.
A neural network model trained with Connectionist Temporal
Classification (CTC) \citep{Graves2006} achieved promising results on the Wall Street Journal
corpus \citep{Graves2014, Hannun2014a}. A similar setup was used to obtain state-of-the-art
results on the Switchboard task as well \citep{Hannun2014b}.
Both of these models were trained to predict sequences of characters and were
later combined with a word level language model. Furthermore, when the
language model was implemented as a CTC-specific Weighted Finite State
Transducer, decoding accuracies competitive with DNN-HMM hybrids were
obtained \citep{Miao2015}.

At the same time, a new direction of neural network research has emerged that
deals with models that learn to focus their ``attention'' to specific parts
of their input. Systems of this type have
shown very promising results on a variety of tasks including machine translation
\citep{Bahdanau2014}, caption generation \citep{Xu2015}, handwriting synthesis
\citep{Graves2013}, visual object classification \citep{Mnih2014} and phoneme
recognition \citep{Chorowski2014,Chorowski2015}.

In this work, we investigate the application of an Attention-based
Recurrent Sequence Generator (ARSG) as a part of an end-to-end 
LVCSR system. We start from the system proposed in \citep{Chorowski2015} 
and make the following contributions:
\begin{enumerate}
\item We show how training on long sequences can be made
    feasible by limiting the area explored by the attention
    to a range of most promising locations. This reduces the
    total training complexity from quadratic to linear,
    largely solving the scalability issue of the approach.
    This has already been proposed \citep{Chorowski2015}
    under the name ``windowing'', but was used only at the
    decoding stage in that work. 
\item In the spirit of he Clockwork RNN \citep{Koutnik2014} and
  hierarchical gating RNN \citep{Chung2015}, we introduce a recurrent
  architecture that successively reduces source sequence length by
  pooling frames neighboring in time. \footnote{This
      mechanism has been 
     recently independently proposed in \citep{Chan2015}.}
\item We show how a character-level ARSG
      and $n-$gram word-level language model can be combined
      into a complete system using  the
      Weighted Finite State Transducers (WFST) framework.
\end{enumerate}

\section{Attention-based Recurrent Sequence Generators for Speech}

The system we propose is a neural network that can map sequences of speech
frames to sequences of characters. While the whole system is differentiable and
can be trained directly to perform the task at hand, it can still be divided
into different functional parts that work together to learn how to \emph{encode}
the speech signal into a suitable feature representation and to \emph{decode}
this representation into a sequence of characters.
We used RNNs for both the encoder and decoder\footnote{The word ``decoder'' 
    refers to a network in this context, not to the final
    recognition algorithm.}
parts of the system. 
The decoder combines
an RNN and an attention mechanism into an Attention-based Recurrent Sequence
Generator that is able to learn the alignment between its input and its output.
Therefore, we will first discuss RNNs, and subsequently, how they can be combined
with attention mechanisms to perform sequence alignment.

\subsection{Recurrent Neural Networks}
\label{sec:recurrent}

There has been quite some research into Recurrent Neural Networks (RNNs) for
speech recognition \citep{robinson1996use,lippmann1989review} and this can
probably be explained to a large extent by the elegant way in which they can
deal with sequences of variable length.

Given a sequence of feature vectors $(\mathbf{x}_{1},\cdots,\mathbf{x}_{T})$,
a standard RNN computes a corresponding sequence of hidden state vectors
$(\mathbf{h}_{1},\cdots,\mathbf{h}_{T})$ using
\begin{equation}
\mathbf{h}_{t} = g(\mathbf{W}_{xh}\mathbf{x}_{t} +
\mathbf{W}_{hh}\mathbf{h}_{t-1} + \mathbf{b}_{h}),
\label{eq:rnn}
\end{equation}
where $\mathbf{W}_{xh}$ and $\mathbf{W}_{hh}$ are matrices of trainable
parameters that represent the connection weights of the network and
$\mathbf{b}_{h}$ is a vector of trainable bias parameters.
The function $g(\cdot)$ is often a non-linear squashing function
like the hyperbolic tangent and applied element-wise to its input.
The hidden states can be used as features that serve as inputs to a layer that
performs a task like classification or regression. Given that this output layer and
the objective to optimize are differentiable, the gradient of this objective
with respect to the parameters of the network can be computed with
backpropagation through time.
Like feed-forward networks, RNNs can process discrete input data by
representing it as 1-hot-coding feature vectors.

An RNN can be used as a statistical model over sequences of
labels. For that, it is trained it to
predict the probability of the next label conditioned on the part of the
sequence it has already processed. If $(y_{1},\cdots,y_{T})$ is a sequence of
labels, an RNN can be trained to provide the conditional distribution the next label using
\begin{align*}
p(y_{t}|y_{1},\cdots,y_{t-1})&=p(y_{t}|\mathbf{h}_{t})\\
                              &=\text{softmax}(\mathbf{W}_{hl}\mathbf{h}_{t} + \mathbf{b}_{l}),
\end{align*}
where $\mathbf{W}_{hl}$ is a matrix of trainable connection weights,
$\mathbf{b}_l$ is a vector of bias parameters and
$\text{softmax}_i(\mathbf{a})=\frac{\exp(a_i)}{\sum_j\exp(a_j)}$.
The likelihood of the complete sequence is now given by
$p(y_1)\prod_{t=2}^{T}p(y_t|y_{1},\cdots,y_{t-1})$.
This distribution can be used to \emph{generate} sequences by either sampling
from the distribution $p(y_{t}|y_{1},\cdots,y_{t-1})$ or choosing the most likely labels
iteratively.  

Equation \ref{eq:rnn} defines the simplest RNN, however in
practice usually more advanced equations define the
dependency of $\mathbf{h}_{t}$
on $\mathbf{h}_{t-1}$. Famous examples of these so-called
recurrent transitions are Long Short Term Memory
\citep{Hochreiter1997} and Gated Recurrent Units (GRU)
\citep{Cho2014}, which are both designed to better handle
long-term dependencies. In this work we use GRU for 
it has a simpler architecture and is easier to implement
efficiently. The hidden states $\mathbf{h}_t$ are computed
using the following equations:
\begin{align*}
    \mathbf{z}_t &= \sigma(\mathbf{W}_{xz} \mathbf{x}_t + \mathbf{U}_{hz} \mathbf{h}_{t-1}), \\
    \mathbf{r}_t &= \sigma\left( \mathbf{W}_{xr} \mathbf{x}_t + \mathbf{U}_{hr} \mathbf{h}_{t-1} \right), \\
    \tilde{\mathbf{h}}_t &= \tanh\left( \mathbf{W}_{xh} \mathbf{x}_t + \mathbf{U}_{rh} (\mathbf{r}_t \otimes \mathbf{h}_{t-1}) \right), \\
    \mathbf{h}_t &= (1 - \mathbf{z}_t) \mathbf{h}_{t - 1} + \mathbf{z}_t \tilde{\mathbf{h}}_t, 
\end{align*}
% Dima B.: why not use vector form for these equations?
where $\mathbf{\tilde{h}}_t$ are candidate activations, $\mathbf{z}_t$ and
$\mathbf{r}_t$ are update and reset gates respectively. The
symbol $\otimes$ signifies element-wise multiplication.

To obtain a model that uses information from both future frames and past
frames, one can pass the input data through two recurrent neural networks that
run in opposite directions and concatenate their hidden state vectors.
Recurrent neural network of this type are often referred
to as \emph{bidirectional} RNNs.

\begin{figure}[t]
\centering
\begin{tikzpicture}[->,thick]
\scriptsize
\tikzstyle{main}=[circle, minimum size = 7mm, thick, draw =black!80, node distance = 12mm]
\foreach \name in {1,...,6}
    \node[main, fill = white!100] (l\name) at (\name,2.3) {$\mathbf{h}^2_\name$};
\foreach \name in {1,...,6}
    \node[main, fill = white!100] (k\name) at (\name,3.1) {$\hat{\mathbf{h}}^2_\name$};
\foreach \name in {1,...,6}
    \node[main, fill = white!100] (i\name) at (\name,.8) {$\hat{\mathbf{h}}^1_\name$};
\foreach \name in {1,...,6}
    \node[main, fill = white!100] (h\name) at (\name,0) {$\mathbf{h}^1_\name$};
\foreach \name in {1,...,6}
    \node[main, fill = white!100] (x\name) at (\name,-1.5) {$\mathbf{x}_\name$};
\foreach \h in {1,...,6}
       {
        \path (x\h) edge [bend right] (i\h);
        \path (x\h) edge (h\h);
        \path (i\h) edge [bend left] (k\h);
        \path (h\h) edge [bend right] (l\h);
        \path (h\h) edge [bend left] (k\h);
        \path (i\h) edge (l\h);
       }
\foreach \current/\next in {1/2,2/3,3/4,4/5,5/6} 
       {
        \path (i\next) edge (i\current);
        \path (h\current) edge (h\next);
        \path (k\next) edge (k\current);
        \path (l\current) edge (l\next);
       }
    %\node[main] (G-\name) at (\x,0) {$\name$};
\end{tikzpicture}
\caption{Two Bidirectional Recurrent Neural Networks stacked on top of each
other.}
\label{fig:birnn}
\end{figure}
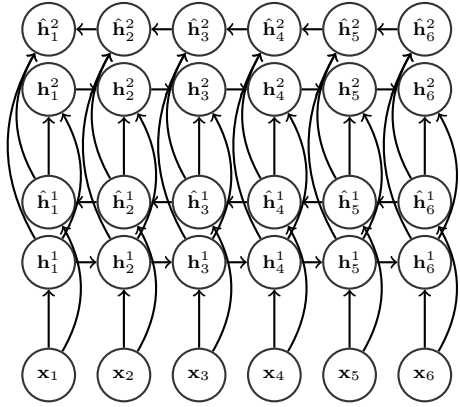

Finally, it has been shown that better results for speech recognition
tasks can be obtained by stacking multiple layers of recurrent neural networks
on top of each other \citep{Graves2013speech}. This can simply be done by
treating the sequence of state vectors
$(\mathbf{h}_1,\cdots,\mathbf{h}_{T})$ as
the input sequence for the next RNN in the pile.
Figure~\ref{fig:birnn} shows an example of two bidirectional RNNs that have been
stacked on top of each other to construct a deep architecture.

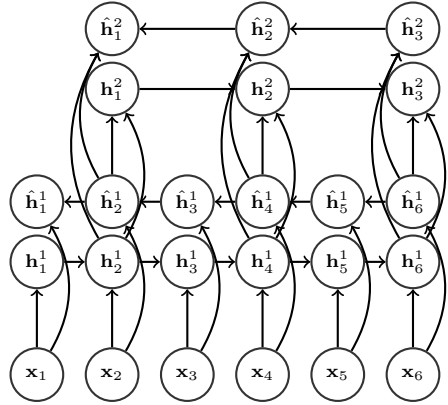
\begin{figure}[t]
\centering
\begin{tikzpicture}[->,thick]
\scriptsize
\tikzstyle{main}=[circle, minimum size = 7mm, thick, draw =black!80, node distance = 12mm]
\foreach \name in {1,...,3}
    \node[main, fill = white!100] (l\name) at (\name*2,2.3) {$\mathbf{h}^2_\name$};
\foreach \name in {1,...,3}
    \node[main, fill = white!100] (k\name) at (\name*2,3.1) {$\hat{\mathbf{h}}^2_\name$};
\foreach \name in {1,...,6}
    \node[main, fill = white!100] (i\name) at (\name,.8) {$\hat{\mathbf{h}}^1_\name$};
\foreach \name in {1,...,6}
    \node[main, fill = white!100] (h\name) at (\name,0) {$\mathbf{h}^1_\name$};
\foreach \name in {1,...,6}
    \node[main, fill = white!100] (x\name) at (\name,-1.5) {$\mathbf{x}_\name$};
\foreach \h in {1,...,6}
        {
         \path (x\h) edge [bend right] (i\h);
         \path (x\h) edge (h\h);
        }
\foreach \bot/\top in {2/1,4/2,6/3}
        {
         \path (i\bot) edge [bend left] (k\top);
         \path (i\bot) edge (l\top);
         \path (h\bot) edge [bend right] (l\top);
         \path (h\bot) edge [bend left] (k\top);
        }

\foreach \current/\next in {1/2,2/3,3/4,4/5,5/6} 
       {
        \path (i\next) edge (i\current);
        \path (h\current) edge (h\next);
       } 
\foreach \current/\next in {1/2,2/3}
       {
        \path (k\next) edge (k\current);
        \path (l\current) edge (l\next);
       }
    %\node[main] (G-\name) at (\x,0) {$\name$};
\end{tikzpicture}
\caption{A pooling over time BiRNN: the upper layer runs twice slower
  then the lower one. It can average, or subsample (as shown in the
  figure) the hidden states of the layer below it.}
\label{fig:pooling_birnn}
\end{figure}

\subsection{Encoder-Decoder Architecture}
\label{sec:encoder-decoder}
Many challenging tasks involve inputs and outputs which may have variable length.
Examples are machine translation and speech recognition, where both input and 
output have variable length; and image caption generation, where the captions may 
have variable lengths. 

Encoder-decoder networks are often used to deal with variable
length input and output sequences \citep{Cho2014,Sutskver2014}. The
encoder is a network that transforms the input into an
intermediate representation. The decoder is typically an RNN
that uses this representation in order to generate the
outputs sequences as described in \ref{sec:recurrent}. 

In this work, we use a deep BiRNN as an encoder. Thus, the
representation is a sequence of BiRNN state vectors
$(\mathbf{h_1},\ldots,\mathbf{h_L})$. For a standard deep
BiRNN, the sequence $(\mathbf{h_1},\ldots,\mathbf{h_L})$ is
as long as the input of the bottom-most layer, which in the
context of speech recongnition means one $\mathbf{h}_i$ for
every 10ms of the recordings. We found that for our
decoder (see \ref{sec:attention}) such representation is
overly precise and contains much redundant information. 
This led us to add pooling between BiRNN
layers as illustrated by Figure \ref{fig:pooling_birnn}.

\subsection{Attention-equipped RNNs}
\label{sec:attention}

The decoder network in our system is an Attention-based
Recurrent Sequence Generator (ARSG). This subsection
introduces ARSGs and explains the motivation behind our
choice of an ARSG for this study.

While RNNs can process and generate sequential data, the
length of the sequence of hidden state vectors is always
equal to the length of the input sequence.  One can aim to
learn the alignment between these two sequences to model a
distribution
$p(y_{1},\cdots,y_{T}|\mathbf{h}_1,\cdots,\mathbf{h}_{L})$
for which there is no clear functional dependency between
$T$ and $L$.

\begin{figure}[t]
\centering
  \includegraphics[width=7.5cm]{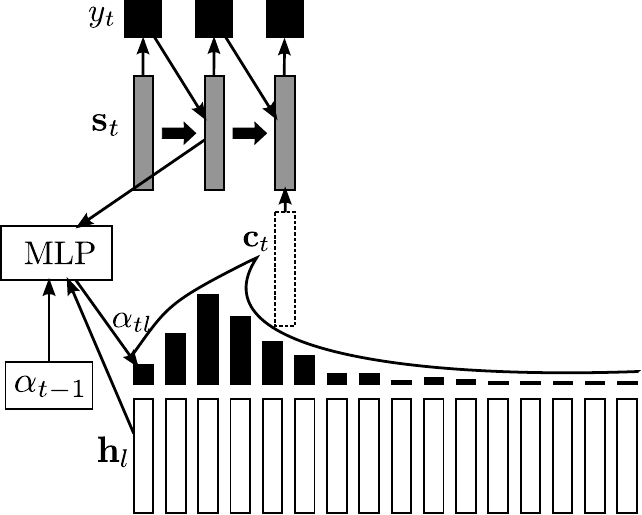}
  \caption{Schematic representation of the Attention-based Recurrent Sequence Generator.
At each time step $t$, an MLP combines the hidden state
$\mathbf{s}_{t-1}$ with all
the input vectors $\mathbf{h}_l$ to compute the attention weights $\alpha_{tl}$.
Subsequently, the new hidden state $\mathbf{s}_t$ and prediction for output
label $y_t$ can be computed.}
\label{fig:arsg}
\end{figure}

An ARSG produces an output sequence $(y_1,\cdots,y_T)$ one element
at a time, simultaneously aligning each generated element to
the encoded input sequence
$(\mathbf{h}_{1},\cdots,\mathbf{h}_L)$.  It is composed
of an RNN and an additional subnetwork called `attention
mechanism'. The attention selects the temporal locations over
the input sequence that should be used to update the hidden
state of the RNN and to make a prediction for the next
output value.
Typically, the selection of elements from the input sequence
is a weighted sum $\mathbf{c}_t=\sum_l \alpha_{tl}\mathbf{h}_l$, where
$\alpha_{tl}$ are called the attention weights and we require that
$\alpha_{tl} \geq 0$ and that $\sum_l \alpha_{tl}=1$.
See Figure~\ref{fig:arsg} for a schematic representation of an ARSG.

The attention mechanism used in this work is an improved
version of the hybrid attention with convolutional features
from \citep{Chorowski2015}, which is described by the
following equations:
\begin{align}
    \mathbf{F}= \mathbf{Q} * \boldsymbol{\alpha}_{t-1} 
    \label{eq:att1}\\
    e_{tl} = \mathbf{w}^\top \tanh(\mathbf{W} \mathbf{s}_{t-1} + 
    \mathbf{V} \mathbf{h}_l + 
    \mathbf{U} \mathbf{f}_{l} +
    \mathbf{b}) 
    \label{eq:att2}\\
    \alpha_{tl} = 
        \exp(e_{tl}) \left/
        \sum\limits_{l=1}^L \exp(e_{tl}) \right..
    \label{eq:att3}        
\end{align}
where $\mathbf{W}$, $\mathbf{V}$, $\mathbf{U}$, $\mathbf{Q}$ are parameter
matrices, $\mathbf{w}$ and $\mathbf{b}$ are parameter
vectors, $*$ denotes convolution, $\mathbf{s}_{t-1}$ stands for the
previous state of the RNN component of the ARSG. We explain how it works
starting from the end: \eqref{eq:att3} shows how the weights $\alpha_{tl}$ 
are obtained by normalizing the scores $e_{tl}$. As
illustrated by \eqref{eq:att2}, the score depends
on the previous state $\mathbf{s}_{t-1}$, the content in the
respective location $\mathbf{h}_l$ and the vector of so-called
convolutional features $\mathbf{f}_l$. The name
``convolutional''  comes from the
convolution along the time axis used in 
\eqref{eq:att1} to compute the
matrix $\mathbf{F}$ that comprises all feature vectors 
$\mathbf{f}_l$. 

Simply put, the attention mechanism described above combines
information from three sources to decide where to focus at
the step $t$: the decoding history contained in $\mathbf{s}_{t-1}$, 
the content in the candidate location $\mathbf{h}_l$ and the
focus from the previous step described by attention weights
$\boldsymbol{\alpha_{t-1}}$. It is shown in
\citep{Chorowski2015} that making the attention
location-aware, that is using  $\boldsymbol{\alpha}_{t-1}$
in the equations defining $\boldsymbol{\alpha}_{t}$, is crucial 
for reliable behaviour on long input sequences.

A disadvantage of the approach from \citep{Chorowski2015} is
the complexity of the training procedure, which is $O(LT)$
since weights $\alpha_{tl}$ have to be computed for all
pairs of input and output positions. The same paper
showcases a windowing approach that reduces the complexity
of decoding to $O(L + T)$. In this work we apply the windowing
at the training stage as well. Namely, we constrain the
attention mechanism to only consider positions from the
range $(m_{t-1} - w_l, \ldots, m_{t-1} + w_r)$, where $m_{t-1}$ is the
median of $\boldsymbol{\alpha}_{t-1}$, interpreted in this context as a
distribution. The values $w_l$ and $w_r$ define how much the window
expands to the left and to the right respectively. 
This modification makes training significantly faster.

Apart from the speedup it brings, windowing can be also very
helpful for starting the training procedure. From our
experience, it becomes increasingly harder to train an ARSG
completely from scratch on longer input sequences. We found
that providing a very rough estimate of the desired
alignment at the early training stage is an effective way to
quickly bring network parameters in a good range.
Specifically, we forced the network to only choose from
positions in the range $R_t = (s_{min} + tv_{min}, \ldots,
s_{max} + tv_{max})$. The numbers $s_{min}$, $s_{max}$,
$v_{min}$, $v_{max}$ were roughly estimated from the training
set so that the number of leading silent frames for training
utterances were between $s_{min}$ and $s_{max}$ and so that
the speaker speed, i.e. the ratio between the transcript and the
encoded input lengths, were between $v_{min}$ and $v_{max}$.
We aimed to make the windows $R_t$ as narrow as possible,
while keeping the invariant that the character $y_t$ was
pronounced within the window $R_t$. The resulting sequence
of windows is quickly expanding, but still it was sufficient
to quickly move the network out of the random initial mode, in which
it had often aligned all characters to a single location in
the audio data. We note, that the median-centered windowing
could not be used for this purpose, since it relies on the
quality of the previous alignment to define the window for
the new one.

\section{Integration with a Language Model}
\label{sec:fsts}

Although an ARSG by construction implicitly learns how an
output symbol depends on the previous ones, the transcriptions of the
training utterances are typically insufficient to learn
a high-quality language model. For this reason, we
investigate how an ARSG can be integrated with a language
model. The main challenge is that in speech recognition
word-based language models are used, whereas our ARSG models
a distribution over character sequences. 

We use the Weighted Finite State Transducer (WFST) framework 
\citep{Mohri2002,Allauzen2007} 
to build a character-level language model from a word-level
one. A WFST is a finite automaton, whose transitions have
weight and input and output labels. It defines a cost of
transducing an input sequence into an output sequence by
considering all pathes with corresponding sequences of input
and output labels. The composition operation can be used to
combine FSTs that define different levels of representation,
such as characters and words in our case.

We compose the language model Finite State Transducer (FST)
$G$ with a lexicon FST $L$ that simply spells out the
letters of each word. More specifically, we build an FST 
$T= \textrm{min}(\textrm{det}(L\circ G))$ to define the
log-probability for character sequences. We push
the weights of this FST towards the starting state to help
hypothesis pruning during decoding.

For decoding we look for a transcript $y$ that minimizes the
cost $L$ which combines the encoder-decoder (ED) and the
language model (LM) outputs as follows:
\begin{equation}
    \label{eq:joint_cost}
    L = -\log p_{ED}(y|x) - \beta \log p_{LM}(y) - \gamma T 
\end{equation}
where $\beta$ and $\gamma$ are tunable parameters.
The last term $\gamma T$ is important, because without it
the LM component dominates and the cost $L$ is minimized by
too short sequences. We note that the same criterion for
decoding was proposed in \citep{Hannun2014a} for a CTC network.

Integrating an FST and an ARSG in a beam-search decoding is easy
because they share the property that the current state
depends only on the previous one and the input symbol. 
Therefore one can use a simple left-to-right beam
search algorithm similar to the one described in
\citep{Sutskver2014} to approximate the value of $y$ that minimizes $L$.

The determinization of the FST becomes impractical for moderately
large FSTs, such as the trigram model shipped with the Wall
Street Journal corpus (see Subsection \ref{sec:data}). To handle
non-deterministic FSTs we assume that its weights are in the
logarithmic semiring and compute the total log-probability
of all FST paths corresponding to a character prefix from
the beam. This probability can be quickly recomputed when a
new character is added to the prefix.

\section{Related Work}
A popular method to train networks to perform sequence prediction
is Connectionist Temporal Classification 
\citep{Graves2006}.
 It has been used with
great success for both phoneme recognition \citep{Graves2013speech} and character-based LVCSR
\citep{Graves2014,Hannun2014a,Hannun2014b,Miao2015}.
CTC allows recurrent neural networks to predict sequences that are shorter than
the input sequence by summing over all possible alignments between the output
sequence and the input of the CTC module. This summation is done using dynamic
programming in a way that is similar to the forward and backward passes that are
used to do inference in an HMM. In the CTC approach,
output labels are conditionally independent given the
alignment and the output sequences. In the context of speech
recognition, this means that a CTC
network lacks a language model,
which greatly boosts the system performance when added to a
trained CTC network \citep{Hannun2014a, Miao2015}.

An extension of CTC is the RNN Transducer which combines two RNNs into a
sequence transduction system \citep{graves2012sequence,boulanger2013high}.
One network is similar to a CTC network and runs at the
same time-scale as the input sequence, while a separate RNN models the
probability of the next label output label conditioned on the previous ones.
Like in CTC, inference is done with a dynamic programming method similar to the
backward-forward algorithm for HMMs, but taking into account the constraints
defined by both of the RNNs. Unlike CTC, 
RNN transduction systems can also generate output sequences that are
longer than the input. RNN Transducers have
led to state-of-the-art results in phoneme recognition on the TIMIT
dataset \citep{Graves2013speech} which were recently matched by an ASRG
network \citep{Chorowski2015}.

The RNN Transducer and ARSG approaches are roughly
equivalent in their capabilities. In both approaches
an implicit language model is learnt jointly with the rest
of the network. The main difference between the approaches 
is that in ARSG the alignment is explicitly computed by the
network, as opposed to dealing with a distribution of
alignments in the RNN Transducer. We hypothesize that this
difference might have a major impact on the further development 
of these methods.

Finally, we must mention two very recently published works
that partially overlap with the content of this paper.
In \citep{Chan2015} Encoder-Decoder for
character-based recognition, with the model being quite
similar to ours. In particular, in this work 
pooling between the BiRNN layers is also proposed.
Also, in \citep{Miao2015} using FSTs to build 
a character-level model from an n-gram model is advocated.
We note, that the research described in this paper was
carried independently and without communication with the
authors of both aforementioned works.
\section{Experiments}
\label{sec:exp}

\subsection{Data}
\label{sec:data}

We trained and evaluated our models \footnote{Our code is available at
    https://github.com/rizar/attention-lvcsr}
on the Wall Street Journal (WSJ) corpus
(available at the Linguistic Data Consortium as LDC93S6B and LDC94S13B).
Training was done on the 81 hour long SI-284 set of about 37K sentences.
As input features, we used 40 mel-scale filterbank coefficients together with the
energy. These 41 dimensional features were extended with their first and second
order temporal derivatives to obtain a total of 123 feature values per frame.
Evaluation was done on the ``eval92'' evaluation set.
Hyperparameter selection was performed on the ``dev93'' set. 
For language model integration we
have used the 20K closed
vocabulary setup and the bigram and trigram language model
that were provided with the data
set. We use the same text preprocessing as in
\citep{Hannun2014a}, leaving only 32 distinct labels:
26 characters, apostrophe, period, dash, space, noise and
end-of-sequence tokens.

\subsection{Training}

Our model used $4$ layers of 250 forward + 250 backward
 GRU units in the encoder, with the top two
layers reading every second of hidden states of the network
below it (see Figure \ref{fig:pooling_birnn}). Therefore, the encoder reduced the utterance length
by the factor of 4. A centered convolution filter of width 200
was used in the attention mechanism to extract a single
feature from the previous step alignment as described in
\eqref{eq:att3}.

The AdaDelta algorithm \citep{Zeiler2012} with gradient clipping was used
for optimization. We initialized all the weights randomly from an
isotropic Gaussian distribution with variance
$0.1$.

We used a rough estimate of the proper alignment for the
first training epoch as described in Section \ref{sec:attention}.
After that the training was restarted with the windowing
described in the same section. The window parameters were
$w_L=w_R=100$, which corresponds to considering a large 8 second
long span of audio data at each step, taking into account the pooling done
between layers.  Training with the
AdaDelta hyperparameters $\rho=0.95$, $\epsilon=10^{-8}$ was
continued until log-likelihood stopped improving. Finally,
we annealed the best model in terms of CER by
restarting the training with $\epsilon=10^{-10}$.

We found regularization necessary for the best performance.
The column norm constraint 1 was imposed on all weight
matrices \citep{Hinton2012}. This corresponds to constraining
the norm of the weights of all the connections incoming to a unit.

\subsection{Decoding and Evaluation}

As explained in Section $\ref{sec:fsts}$, we used beam search to
minimize the combined cost $L$ defined by
\eqref{eq:joint_cost}.  We finished when $k$ terminated
sequences cheaper than any non-terminated sequence in the
beam were found. A sequence was considered terminated when
it ended with the special end-of-sequence token,
which the network was trained to generate in the end of each
transcript.

To measure the best performance we used
the beam size 200, however this brought us only $\approx 10\%$
relative improvement over beam size 10. We used parameter
settings $\alpha=0.5$ and $\gamma=1$ with a language model and
$\gamma=0.1$ without one. It was necessary to use an asymmetric
window for the attention when decoding with large $\gamma$.
More specifically, we reduced $w_L$ to 10.
Without this trick, the cost $L$ could be infinitely minimized
by looping across the input utterance, for the penalty for jumping back in time
included in $\log p(y|x)$  was not high enough.

\subsection{Results}
\label{sec:res}
Results of our experiments are gathered in Table
\ref{tab:results}. Our model shows performance superior to
that of CTC systems when no external language model is used. 
The improvement from adding an external language model is
however much larger for CTC-based systems. The final
peformance of our model is better than the one reported in
\citep{Hannun2014a} (13.0\% vs 14.1\%), but worse than the the one
from \citep{Miao2015} (11.3\% vs 9.0\%) when the same
language models are used.

\begin{table}
\label{tab:results}
\caption{Character Error Rate (CER) and Word Error Rate (WER) scores for our
setup on the Wall Street Journal Corpus in comparison with other results from
the literature. Note that our results are not directly
comparable with those of networks predicting phonemes instead
of characters, since phonemes are easier targets.}%
% Results from https://docs.google.com/spreadsheets/d/14Gm_gjMzp_7rcVGSIR0jaV4Fdogwm2UBK1CeQ87Hv78/edit#gid=1483356118
%
\begin{tabular}{l c c}
Model &CER &WER\\
\hline
Encoder-Decoder & 6.4 &18.6\\
Encoder-Decoder + bigram LM & 5.3 & 11.7\\
Encoder-Decoder + trigram LM & 4.8 & 10.8\\
Encoder-Decoder + extended trigram LM & 3.9 & 9.3\\
\hline
Graves and Jaitly (2014)\\
    \hspace{0.5cm} CTC &9.2 &30.1\\
    \hspace{0.5cm} CTC, expected transcription loss &8.4 &27.3\\
\hline
Hannun et al. (2014)\\
    \hspace{0.5cm} CTC &10.0 &35.8\\
    \hspace{0.5cm} CTC + bigram LM &5.7 &14.1\\
\hline 
Miao et al. (2015), \\ \hspace{0.5cm} CTC for phonemes + lexicon  &- & 26.9\\
 \hspace{0.5cm} CTC for phonemes + trigram LM &- & 7.3\\
 \hspace{0.5cm} CTC + trigram LM &- & 9.0
\end{tabular}
\end{table}

\section{Discussion}

A major difference between the CTC and ARSG approaches is
that a language model is implicitly learnt in the latter.
Indeed, one can see that an RNN sequence model as explained
in \ref{sec:recurrent} is literally contained in an ARSG as a subnetwork.
We believe that this is the reason for the greater
performance of the ARSG-based system when no external LM is
used. However, this implicit language model was trained on a
relatively small corpus of WSJ transcripts containing less
than 4 million characters. It has been reported that
RNNs overfit on corpora of
such size \citep{Graves2013} and in our experiments 
we had to combat overfitting as well. Using the weight
clipping brought a consistent performance
gain but did not change the big picture.
For these reasons, we hypothesize that overfitting of the
internal RNN language model was one of the main reasons why
our model did not reach the performance level reported in
\citep{Miao2015}, where a CTC network is used.

That being said, we treat it as an advantage of the
ARSG that it supports joint training of a language model
with the rest of the network. For one, WSJ contains
approximately only 80
hours of training data, and overfitting might be less of an
issue for corpora containing hundreds or even thousands
hours of annotated speech. 
For two, an RNN language model
trained on a large text corpus could be integrated in an
ARSG from the beginning of training by using the states of
this language model as an additional input of the ARSG. 
We suppose that this would block the incentive of memorizing 
the training utterances, and thereby reduce the overfitting.
In addition, no extra n-gram model would be required.
We note that a similar idea has been already proposed in
\citep{gulcehre2015using} for machine translation. 

Finally, trainable integration with an n-gram language model
could also be investigated.

\subsection{Conclusion}
In this work we showed how an Encoder-Decoder network with an
attention mechanism can be used to build a LVCSR system.
The resulting approach is significantly simpler than the
dominating HMM-DNN one, with fewer training stages, much less
auxiliary data and less domain expertise involved. Combined with
a trigram language model our system shows decent, although
not yet state-of-the-art performance.

We present two methods to improve the computational
complexity of the investigated model. First, we propose pooling
over time between BiRNN layers to reduce the length of the
encoded input sequence. Second, we propose to use windowing
during training to ensure that the decoder network performs a
constant number of operations for each output character.
Together these two methods facilitate application of
attention-based models to large-scale speech recognition.

Unlike CTC networks, our model has an intrinsic
language-modeling capability. Furthermore, it has a potential
to be trained jointly with an external language model. 
Investigations in this direction are likely to be a part of
our future work.

\subsubsection*{Acknowledgments}
The experiments were conducted using Theano
\citep{bergstra+al:2010-scipy,Bastien-Theano-2012}, 
Blocks and Fuel \citep{vanmerrienboer_blocks_2015} libraries.

The authors would like to acknowledge the support of the following agencies for
research funding and computing support: National Science Center (Poland), 
NSERC, Calcul Qu\'{e}bec, Compute Canada,
the Canada Research Chairs and CIFAR. Bahdanau also thanks Planet
Intelligent Systems GmbH and Yandex.

\bibliographystyle{apalike}
\bibliography{paperrefs}

\begin{thebibliography}{}

\bibitem[Allauzen et~al., 2007]{Allauzen2007}
Allauzen, C., Riley, M., Schalkwyk, J., Skut, W., and Mohri, M. (2007).
\newblock {OpenFst}: A general and efficient weighted finite-state transducer
  library.
\newblock In Holub, J. and {\v Z}{\v d}{\'a}rek, J., editors, {\em
  Implementation and Application of Automata}, number 4783 in Lecture Notes in
  Computer Science, pages 11--23. Springer Berlin Heidelberg.

\bibitem[Bahdanau et~al., 2015]{Bahdanau2014}
Bahdanau, D., Cho, K., and Bengio, Y. (2015).
\newblock Neural machine translation by jointly learning to align and
  translate.
\newblock In {\em International Conference on Learning Representations}.

\bibitem[Bastien et~al., 2012]{Bastien-Theano-2012}
Bastien, F., Lamblin, P., Pascanu, R., Bergstra, J., Goodfellow, I.~J.,
  Bergeron, A., Bouchard, N., and Bengio, Y. (2012).
\newblock Theano: new features and speed improvements.
\newblock Deep Learning and Unsupervised Feature Learning NIPS Workshop.

\bibitem[Bergstra et~al., 2010]{bergstra+al:2010-scipy}
Bergstra, J., Breuleux, O., Bastien, F., Lamblin, P., Pascanu, R., Desjardins,
  G., Turian, J., Warde-Farley, D., and Bengio, Y. (2010).
\newblock Theano: a {CPU} and {GPU} math expression compiler.
\newblock In {\em Proceedings of the Python for Scientific Computing Conference
  ({SciPy})}.

\bibitem[Boulanger-Lewandowski et~al., 2013]{boulanger2013high}
Boulanger-Lewandowski, N., Bengio, Y., and Vincent, P. (2013).
\newblock High-dimensional sequence transduction.
\newblock In {\em ICASSP}, pages 3178--3182. IEEE.

\bibitem[Chan et~al., 2015]{Chan2015}
Chan, W., Jaitly, N., Le, Q.~V., and Vinyals, O. (2015).
\newblock Listen, attend and spell.
\newblock {\em {arXiv}:1508.01211 {[}cs, stat]}.

\bibitem[Cho et~al., 2014]{Cho2014}
Cho, K., van Merrienboer, B., Gulcehre, C., Bougares, F., Schwenk, H., and
  Bengio, Y. (2014).
\newblock Learning phrase representations using {RNN} encoder-decoder for
  statistical machine translation.
\newblock In {\em Empirical Methods of Natural Language Processing}.

\bibitem[Chorowski et~al., 2014]{Chorowski2014}
Chorowski, J., Bahdanau, D., Cho, K., and Bengio, Y. (2014).
\newblock End-to-end continuous speech recognition using attention-based
  recurrent {NN}: First results.
\newblock {\em {arXiv}:1412.1602 {[}cs, stat{]}}.

\bibitem[Chorowski et~al., 2015]{Chorowski2015}
Chorowski, J., Bahdanau, D., Serdyuk, D., Cho, K., and Bengio, Y. (2015).
\newblock Attention-based models for speech recognition.
\newblock In {\em NIPS}.
\newblock to appear.

\bibitem[Chung et~al., 2015]{Chung2015}
Chung, J., Gulcehre, C., Cho, K., and Bengio, Y. (2015).
\newblock Gated feedback recurrent neural networks.
\newblock In {\em ICML-15}.

\bibitem[Graves, 2012]{graves2012sequence}
Graves, A. (2012).
\newblock Sequence transduction with recurrent neural networks.
\newblock {\em arXiv preprint arXiv:1211.3711}.

\bibitem[Graves, 2013]{Graves2013}
Graves, A. (2013).
\newblock Generating sequences with recurrent neural networks.
\newblock {\em {arXiv}:1308.0850}.

\bibitem[Graves et~al., 2006]{Graves2006}
Graves, A., Fern{\'a}ndez, S., Gomez, F., and Schmidhuber, J. (2006).
\newblock Connectionist temporal classification: Labelling unsegmented sequence
  data with recurrent neural networks.
\newblock In {\em ICML-06}.

\bibitem[Graves and Jaitly, 2014]{Graves2014}
Graves, A. and Jaitly, N. (2014).
\newblock Towards end-to-end speech recognition with recurrent neural networks.
\newblock In {\em ICML-14}.

\bibitem[Graves et~al., 2013]{Graves2013speech}
Graves, A., Mohamed, A.-r., and Hinton, G. (2013).
\newblock Speech recognition with deep recurrent neural networks.
\newblock In {\em ICASSP}, pages 6645--6649. IEEE.

\bibitem[Gulcehre et~al., 2015]{gulcehre2015using}
Gulcehre, C., Firat, O., Xu, K., Cho, K., Barrault, L., Lin, H.-C., Bougares,
  F., Schwenk, H., and Bengio, Y. (2015).
\newblock On using monolingual corpora in neural machine translation.
\newblock {\em arXiv preprint arXiv:1503.03535}.

\bibitem[Hannun et~al., 2014a]{Hannun2014b}
Hannun, A., Case, C., Casper, J., Catanzaro, B., Diamos, G., Elsen, E.,
  Prenger, R., Satheesh, S., Sengupta, S., Coates, A., et~al. (2014a).
\newblock Deepspeech: Scaling up end-to-end speech recognition.
\newblock {\em arXiv preprint arXiv:1412.5567}.

\bibitem[Hannun et~al., 2014b]{Hannun2014a}
Hannun, A.~Y., Maas, A.~L., Jurafsky, D., and Ng, A.~Y. (2014b).
\newblock First-pass large vocabulary continuous speech recognition using
  bi-directional recurrent dnns.
\newblock {\em arXiv preprint arXiv:1408.2873}.

\bibitem[Hinton et~al., 2012a]{hinton2012deep}
Hinton, G., Deng, L., Yu, D., Dahl, G.~E., Mohamed, A.-r., Jaitly, N., Senior,
  A., Vanhoucke, V., Nguyen, P., Sainath, T.~N., and Kingsbury, B. (2012a).
\newblock Deep neural networks for acoustic modeling in speech recognition: The
  shared views of four research groups.
\newblock {\em Signal Processing Magazine, IEEE}, 29(6):82--97.

\bibitem[Hinton et~al., 2012b]{Hinton2012}
Hinton, G.~E., Srivastava, N., Krizhevsky, A., Sutskever, I., and
  Salakhutdinov, R.~R. (2012b).
\newblock Improving neural networks by preventing co-adaptation of feature
  detectors.
\newblock {\em arXiv preprint arXiv:1207.0580}.

\bibitem[Hochreiter and Schmidhuber, 1997]{Hochreiter1997}
Hochreiter, S. and Schmidhuber, J. (1997).
\newblock Long short-term memory.
\newblock {\em Neural computation}, 9(8):1735--1780.

\bibitem[Koutnik et~al., 2014]{Koutnik2014}
Koutnik, J., Greff, K., Gomez, F., and Schmidhuber, J. (2014).
\newblock A clockwork {RNN}.
\newblock In {\em ICML-14}.

\bibitem[Lippmann, 1989]{lippmann1989review}
Lippmann, R.~P. (1989).
\newblock Review of neural networks for speech recognition.
\newblock {\em Neural computation}, 1(1):1--38.

\bibitem[Miao et~al., 2015]{Miao2015}
Miao, Y., Gowayyed, M., and Metze, F. (2015).
\newblock {EESEN}: End-to-end speech recognition using deep {RNN} models and
  {WFST}-based decoding.
\newblock {\em {arXiv}:1507.08240 {[}cs]}.

\bibitem[Mnih et~al., 2014]{Mnih2014}
Mnih, V., Heess, N., Graves, A., et~al. (2014).
\newblock Recurrent models of visual attention.
\newblock In {\em NIPS}, pages 2204--2212.

\bibitem[Mohri et~al., 2002]{Mohri2002}
Mohri, M., Pereira, F., and Riley, M. (2002).
\newblock Weighted finite-state transducers in speech recognition.
\newblock {\em Computer Speech \& Language}, 16(1):69--88.

\bibitem[Robinson et~al., 1996]{robinson1996use}
Robinson, T., Hochberg, M., and Renals, S. (1996).
\newblock The use of recurrent neural networks in continuous speech
  recognition.
\newblock In {\em Automatic speech and speaker recognition}, pages 233--258.
  Springer.

\bibitem[Sutskever et~al., 2014]{Sutskver2014}
Sutskever, I., Vinyals, O., and Le, Q.~V. (2014).
\newblock Sequence to sequence learning with neural networks.
\newblock In {\em NIPS}.

\bibitem[van Merri{\"e}nboer et~al., 2015]{vanmerrienboer_blocks_2015}
van Merri{\"e}nboer, B., Bahdanau, D., Dumoulin, V., Serdyuk, D., Warde-Farley,
  D., Chorowski, J., and Bengio, Y. (2015).
\newblock Blocks and fuel: Frameworks for deep learning.
\newblock {\em {arXiv}:1506.00619 {[}cs, stat{]}}.

\bibitem[Xu et~al., 2015]{Xu2015}
Xu, K., Ba, J., Kiros, R., Cho, K., Courville, A., Salakhutdinov, R., Zemel,
  R., and Bengio, Y. (2015).
\newblock Show, attend and tell: Neural image caption generation with visual
  attention.
\newblock In {\em ICML-15}.

\bibitem[Zeiler, 2012]{Zeiler2012}
Zeiler, M.~D. (2012).
\newblock Adadelta: An adaptive learning rate method.
\newblock {\em arXiv preprint arXiv:1212.5701}.

\end{thebibliography}

\end{document}